\documentclass[conference]{IEEEtran}
\usepackage{graphicx} 
\usepackage{booktabs,colortbl,xcolor}
\usepackage{multirow}
\usepackage{amsmath}
\usepackage{stfloats}
\usepackage{hyperref}

\def\BibTeX{{\rm B\kern-.05em{\sc i\kern-.025em b}\kern-.08em
    T\kern-.1667em\lower.7ex\hbox{E}\kern-.125emX}}

\begin{document}

\title{Explanatory Interactive Machine Learning for Bias Mitigation in Visual Gender Classification}

\author{\IEEEauthorblockN{Nathanya Queby Satriani}
\IEEEauthorblockA{\textit{Johannes Kepler University Linz}\\
 Linz, Austria \\
 nathanya.satriani@jku.at} 
 \and
 \IEEEauthorblockN{Djordje Slijepčević}
\IEEEauthorblockA{\textit{Institute of Creative Media Technologies} \\
\textit{St. Pölten University of Applied Sciences}\\
 St. Pölten, Austria \\ djordje.slijepcevic@fhstp.ac.at}
 \and
 \IEEEauthorblockN{Markus Schedl}
\IEEEauthorblockA{\textit{Institute of Computational Perception} \\
 \textit{Johannes Kepler University Linz}\\ 
 Linz, Austria \\
 markus.schedl@jku.at}
 \and
 \IEEEauthorblockN{Matthias Zeppelzauer}
\IEEEauthorblockA{\textit{Institute of Creative Media Technologies} \\
 \textit{St. Pölten University of Applied Sciences}\\
 St. Pölten, Austria\\
 matthias.zeppelzauer@fhstp.ac.at}
}



\maketitle

\begin{abstract}

Explanatory interactive learning (XIL) enables users to guide model training in machine learning (ML) by providing feedback on the model’s explanations, thereby helping it to focus on features that are relevant to the prediction from the user's perspective. In this study, we explore the capability of this learning paradigm to mitigate bias and spurious correlations in visual classifiers, specifically in scenarios prone to data bias, such as gender classification. 
We investigate two methodologically different state-of-the-art XIL strategies, i.e., CAIPI and Right for the Right Reasons (RRR), as well as a novel hybrid approach that combines both strategies.
The results are evaluated quantitatively by comparing segmentation masks with explanations generated using Gradient-weighted Class Activation Mapping (GradCAM) and Bounded Logit Attention (BLA). Experimental results demonstrate the effectiveness of these methods in (i) guiding ML models to focus on relevant image features, particularly when CAIPI is used, and (ii) reducing model bias (i.e., balancing the misclassification rates between male and female predictions). Our analysis further supports the potential of XIL methods to improve fairness in gender classifiers. Overall, the increased transparency and fairness obtained by XIL leads to slight performance decreases with an exception being CAIPI, which shows potential to even improve classification accuracy.


\end{abstract}


\section{Introduction}

\begin{figure}[!t]
    \centering
    \includegraphics[width=1\linewidth]{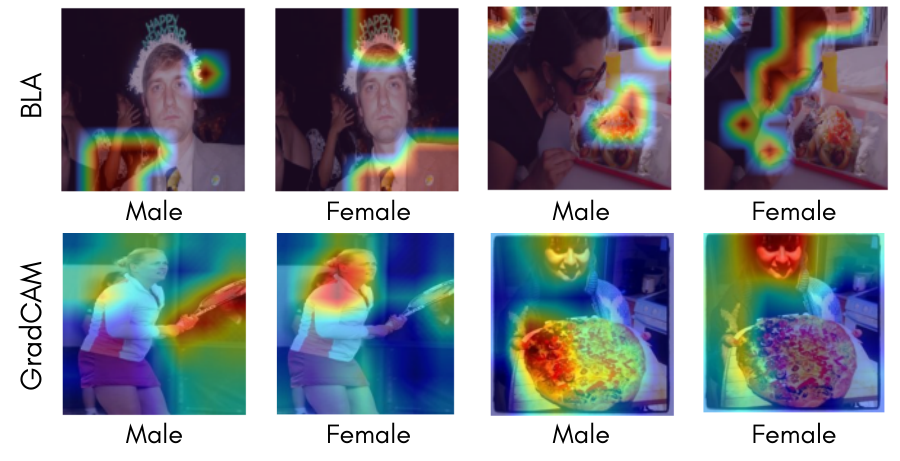}
    \caption{Examples of ambiguous MS~COCO samples on which different pre-trained models disagree. The explanations are presented as attribution maps (Bounded Logit Attention in the top row, GradCAM in the bottom row), which highlight relevant regions for predicting the respective class (below subfigure).}
    \label{fig:bias_examples}
\end{figure}

Data bias is a common phenomenon in Machine Learning (ML) where models learn spurious correlations from the data leading to incorrectly grounded decisions. Data bias is very common in visual classification tasks such as gender classification, where models tend to rely on irrelevant features or stereotypical associations~\cite{oyeniran2022ethical, dong2023probing} (Fig.~\ref{fig:bias_examples} illustrates  examples from our study that exhibit spurious correlations). Various algorithmic approaches have been developed specifically for bias mitigation, either by directly training unbiased models or by removing bias from already trained models~\cite{graziani2023global, schwalbe2024comprehensive, gao2024going, guidotti2018survey}. These methods primarily focus on modifying the training process to limit reliance on biased features. However, another promising  strategy involves leveraging human interaction and feedback during model training. 

Interactive Machine Learning (IML)~\cite{dudley2018} such as Active Learning enables models to iteratively learn from human feedback, dynamically redefining predictions and decision-making. As an extension of IML, Explanatory Interactive Learning (XIL)~\cite{gao2024going,teso2023leveraging} integrates decision explanations derived from explainable AI (XAI)~\cite{adadi2018peeking} methods into the feedback loop. XIL allows users to not only review model predictions but also understand and correct the rationale behind those predictions, enabling more targeted and effective model refinement. Despite its promise, current XAI methods mainly provide one-way transparency, i.e., allowing users to understand model behavior but enabling only very limited means to communicate higher-level information back to the ML model. In the presence of complex biases, such as gender stereotypes, the question arises how effective XIL-based feedback integration strategies are for bias mitigation. This paper tries to answer this question.

In this research, we systematically investigate the capabilities of XIL specifically for detecting and mitigating gender bias in image classifiers. We examine two methodologically different state-of-the-art XIL methods, i.e., CAIPI~\cite{teso2019explanatory} and Right for the Right Reasons (RRR) loss~\cite{ross2017right}, and further propose a novel hybrid approach combining both methods. The main research question we explore is: 
\begin{center}
\textit{To what extent do CAIPI, RRR loss, and their combination mitigate gender bias in image classifiers and guide these models to focus on relevant input features?}
\end{center}
We evaluate this question through a case study on gender classification using a subset of the MS~COCO dataset~\cite{lin2015microsoftcococommonobjects}, employing explainability methods like Gradient-weighted Class Activation Mapping (GradCAM)~\cite{selvaraju2017grad} and Bounded Logit Attention (BLA)~\cite{baumhauer22bounded} to visualize model explanations and iteratively guide learning.

Key contributions of this work are: (i) We investigate the capabilities of XIL to mitigate data bias in image classifiers; 
(ii) We propose a combined strategy integrating CAIPI and RRR to optimize iteratively trained ML models for both accuracy and fairness; 
(iii) We investigate how well the model learns to focus on relevant regions qualitatively using a post-hoc (GradCAM) and self-learned explainability methods (BLA).

\section{Related Work}

This section reviews advancements in three areas relevant to integrating human feedback into ML, namely IML, XAI, and XIL. 
IML focuses on incorporating human input into the model training process, enabling iterative refinement and adaptation to user feedback. Unlike traditional ML where models are trained on static datasets, IML allows users to guide the learning process by correcting errors or providing additional annotations or context, improving both model performance and alignment with human beliefs and expectations~\cite{dudley2018}. A notable approach is Active Learning, which enables models to selectively ask users for annotations on uncertain or highly informative samples, significantly improving learning efficiency and classification performance~\cite{settles2009active}. Relevance Feedback, another common approach, enables iterative refinement based on user-provided relevance assessments, aligning model outputs more closely with user expectations~\cite{Amershi_Cakmak_Knox_Kulesza_2014}. These interactive paradigms are particularly useful in tasks where automated systems tend to struggle with spurious correlations or complex patterns, such as image classification~\cite{teso2023leveraging}.

XAI addresses the ``black-box'' nature of ML models by providing interpretable insights into their decisions~\cite{schwalbe2024comprehensive, gao2024going}. Feature attribution methods identify influential features, enabling the assessment of whether models ground their decisions on relevant information and helping to detect biases~\cite{gunning2019xai}. Most explainability methods are non-interactive and are  applied post hoc to an already trained model, such as Grad-CAM~\cite{selvaraju2017grad}. 
As some post-hoc attribution methods have been criticized for limited faithfulness and robustness, self-explaining XAI methods (e.g., BLA) have been proposed, which provide faithfulness by design.  

XIL combines IML and XAI  by integrating human feedback at the level of model explanations into the interactive learning process~\cite{gao2024going}. XIL enables real-time correction of misclassifications not only through correcting a label like in Active Learning but also by adding feedback on the explanation. XIL  allows the continuous assessment of model behavior while simultaneously adapting it during the training process. 
In the literature, two primary strategies enable XIL approaches: (i) adding an auxiliary loss (e.g., RRR~\cite{ross2017right}, GRADIA~\cite{gao2022aligning}, ProtoPDebug~\cite{bontempelli2023concept}) to penalize the model when it fails to use relevant features, and (ii) employing augmentation techniques to generate counterexamples (e.g., CAIPI~\cite{teso2019explanatory}). 

Comprehensive surveys on XIL offer taxonomies and more detailed insights into these methods~\cite{gao2024going, teso2023leveraging}. Additionally, open-source platforms such as the TrustAI~\cite{slijepvcevic2025trustai} platform support the integration of XIL methods.
While XIL is an extension of IML to more effectively train models in an interactive fashion, its capabilities to detect and mitigate biases represent an open question that we address in this paper.

\section{Methodology}

In the following, we present the proposed methodology and XIL methods employed in our investigation.

\subsection{Overview}

\begin{figure}[b]
    \centering
    \includegraphics[width=1\linewidth]{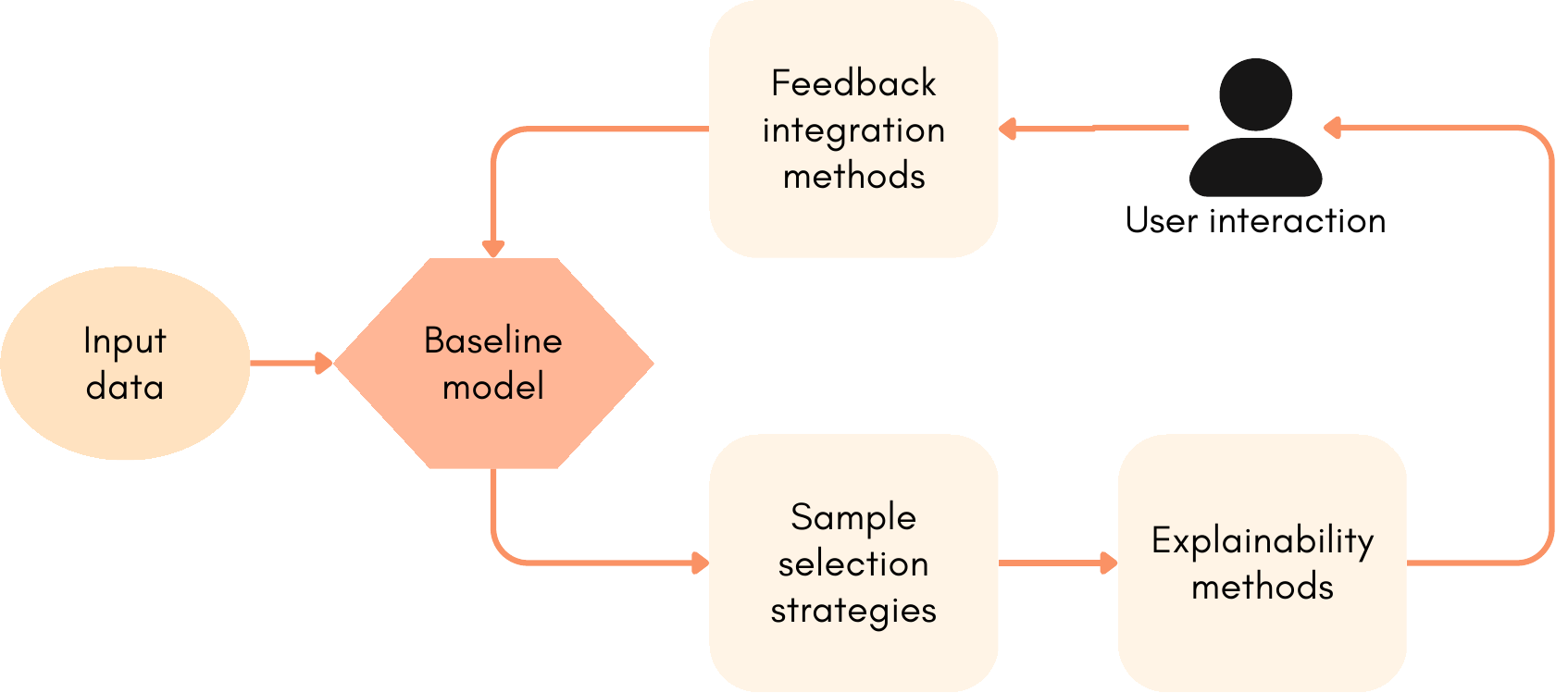}
    \caption{Workflow of our explainability-guided interactive learning setup.}
    \label{fig:abstract_workflow}
\end{figure}

Fig.~\ref{fig:abstract_workflow} presents an overview of the system architecture underlying this work. The process begins with training a classification model on the provided input dataset, which serves as the baseline for subsequent comparisons. A subset of the dataset is  selected using a sample selection strategy (i.e., uncertainty sampling and high-confidence sampling) and serves as the basis for explainability-based interactive learning. 
The classifications of these samples are then explained using GradCAM or BLA. For CAIPI, the ground-truth segmentation masks are used as a baseline to define irrelevant regions, where $k$ counterexamples per sample are generated through image transformations; these augmented samples are then added to the training set. For RRR, the segmentation masks serve as explainability-based ground truth against which the GradCAM or BLA explanations are compared, providing an auxiliary loss (in addition to cross-entropy) to encourage the model to be ``right for the right reasons''~\cite{ross2017right}. The performance of both XIL methods is subsequently compared with the baseline, considering not only classification accuracy but also explainability-based metrics.

\subsection{Baseline Model Architectures}
To establish a baseline model for our comparative analysis, we implemented six convolutional neural networks (CNNs), all pre-trained on the ImageNet dataset: DenseNet121~\cite{DBLP:journals/corr/HuangLW16a}, EfficientNet-B0~\cite{tan2019efficientnet}, GoogLeNet~\cite{szegedy2015going}, MobileNet-V2~\cite{DBLP:journals/corr/abs-1801-04381}, ResNet50~\cite{DBLP:journals/corr/HeZRS15}, and VGG16~\cite{simonyan2014very}. The choice of baseline models was made due to their seamless integration with explainability methods such as GradCAM and BLA, as well as their compatibility with the  RRR loss. Additionally, many recent studies in the field continue to employ these models, making our results more comparable to existing literature. 

\subsection{Sample Selection Strategies}

To rank and select suitable samples from the initial dataset that will serve as the basis for explainability-based interactive learning, we utilized two strategies:

\begin{itemize}
\item \textbf{Uncertainty sampling:} Samples where the model’s predictions were close for at least two different classes, indicating uncertainty of the model.
\item \textbf{High-confidence sampling:} Samples where the model made correct predictions with high confidence ($>90\%$).
\end{itemize}


\subsection{Explainability Methods}

To gain insight into the model’s decision-making and identify potential spurious correlations, we employed two state-of-the-art explainability methods, i.e., GradCAM and BLA. \textbf{GradCAM} highlights relevant regions in the input image by computing the gradients of the target class with respect to the feature maps in the last convolutional layer of a CNN, averaging them, and using them to weight these feature maps~\cite{selvaraju2017grad}. A ReLU function is applied to retain only the positive influence of the features, and these values are then used to generate a heatmap as explanation.

\textbf{BLA} is a trainable module that selects a subset of feature maps in the last convolutional layer via bounding the logits with the $\beta$-function (i.e., inverted ReLU function) to produce inherently learned explanations~\cite{baumhauer22bounded}. Unlike post-hoc methods, BLA is integrated into the model architecture, ensuring the explanation is faithful to the model’s reasoning. Its flexible and modular design supports both training-time and post-hoc integration, and it has shown better alignment with user preferences than GradCAM in prior studies~\cite{baumhauer22bounded}. The BLA module is finetuned in each iteration of the steering process. 

We employed attribution maps generated by these two XAI methods for performance evaluation and to assess whether the model steering led to predictions being based on relevant features instead of spurious correlations.

\subsection{Feedback Integration Methods}

We employed CAIPI and RRR to integrate user feedback into the learning process. Both aim to steer the model away from relying on spurious features and encourage alignment between explanations and human reasoning.

\begin{figure}[t]
    \centering
    \includegraphics[width=1\linewidth]{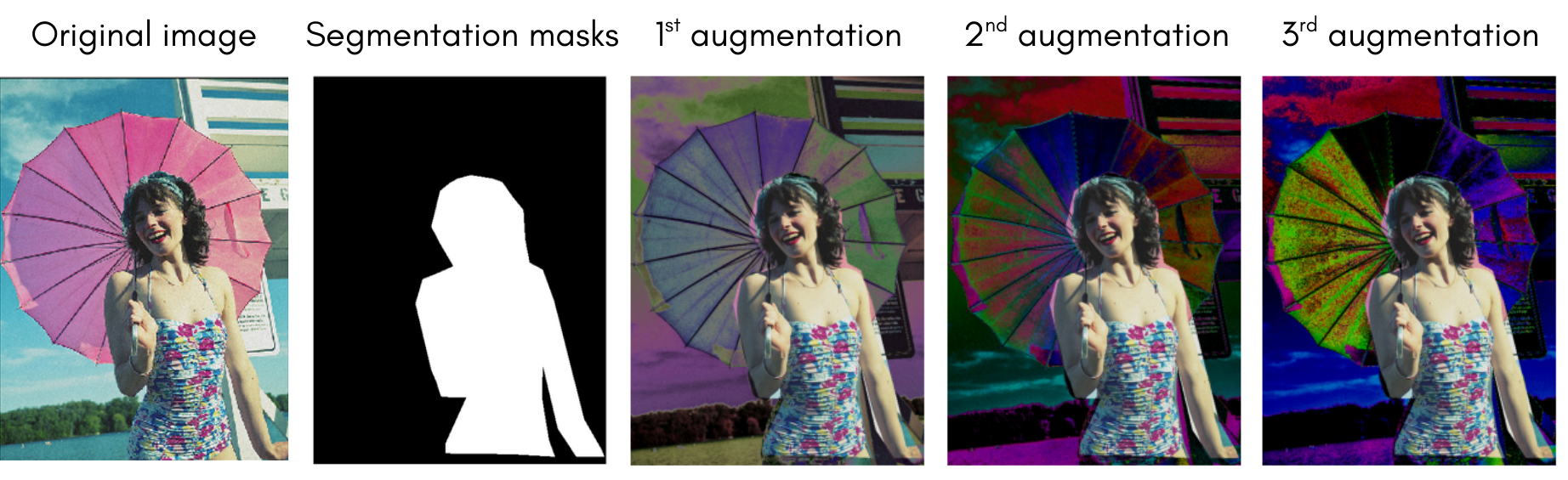}
    \caption{CAIPI data augmentation strategy for \(k=3\), showing different image transformations applied outside the person segmentation mask.}
    \label{fig:caipi}
\end{figure}

\textbf{CAIPI} generates \(k\) synthetic counterexamples by modifying regions marked as irrelevant by the user, while keeping the label unchanged. These countersamples are added to the training set to discourage reliance on the marked features. In our work, we extend the standard CAIPI approach by introducing a fixed sequence of image transformations, i.e., random inversion, posterization, equalization, color jittering, and solarization, to the irrelevant regions based on the ground truth segmentation masks (Fig.~\ref{fig:caipi}). This approach aims to  diversify the input space in irrelevant regions, while preserving semantic meaning in relevant regions. 

\textbf{RRR} guides the learning of the model through an additional term in the loss function of the model, which explicitly penalizes  predictions that rely on regions considered irrelevant by the user. It combines three components: right answer loss, right reasons loss, and weight regularization~\cite{ross2017right}:
\begin{align}
L(\theta, X, y, A) =
&\underbrace{\sum_{n=1}^{N} \sum_{k=1}^{K} -y_{nk} \log(\hat{y}_{nk})}_{\text{Right answers}} \nonumber \\
&+ \lambda_1\underbrace{\sum_{n=1}^{N} \sum_{d=1}^{D}
A_{nd} \left( \frac{\partial}{\partial x_{nd}} \sum_{k=1}^{K} \log(\hat{y}_{nk}) \right)^2}_{\text{Right reasons}} \nonumber \\
&+ \lambda_2\underbrace{\sum_{i} \theta_i^2}_{\text{Regularization}}
\end{align}

The right answers term corresponds to standard cross-entropy loss ensuring correct predictions where \(y_{nk}\) is the ground truth label and \( \hat{y}_{nk} \) is the predicted probability. The right reasons term penalizes large gradients in irrelevant regions, defined by the ground truth mask \( A_{nd} =1 \) for irrelevant features and 0 elsewhere. The L2 regularization term prevents overfitting. The parameters \( \lambda_1 \) and \( \lambda_2 \) serve as weighing factors for the loss regularization terms.

We further propose a \textbf{hybrid approach} composed of CAIPI and RRR. The intuition behind this is that CAIPI and RRR are strongly complementary in how they guide model training: while RRR explicitly guides the training through a loss function, CAIPI implicitly guides the training through data augmentation. Thus, both strategies might benefit from each other in combination. To implement this, we first augment the dataset using CAIPI, then train the model by incorporating the RRR loss function.



\section{Experimental Setup}

This section outlines the experimental configurations used to evaluate the XIL approaches across various settings. We describe the target use case, data preparation, model training procedures, and the evaluation metrics employed to assess performance and bias mitigation.  Code and dataset are available at: \url{https://github.com/fhstp/xil-gender-classification/}. 

\subsection{Use Case Definition}

Gender classification was selected as the primary use case due to its known susceptibility to visual bias and societal implications~\cite{bolukbasi2016mancomputerprogrammerwoman, pmlr-v81-buolamwini18a}. Prior studies have shown that image classifiers often rely on stereotypical or spurious visual cues—such as background objects, clothing, or context, rather than the subject itself~\cite{10.1145/3459990.3460719, dong2023probing}. This makes gender classification a compelling benchmark to investigate how interactive learning guided by explanations can be leveraged to effectively identify and mitigate learned data biases within ML models. 


\subsection{Data Curation}

The dataset used for this study is derived from the MS~COCO dataset, which, however, does not natively provide annotations for gender or sex. Therefore, a manual data selection process was employed to create a binary gender classification dataset\footnote{We are aware that binary gender represents an over-simplification. To limit task complexity and simplify result interpretation, the scope of this study was set to binary classification.}. Although we considered datasets such as PascalVOC~\cite{journals/ijcv/EveringhamGWWZ10} and CelebA~\cite{liu2015faceattributes} for gender classification, we ultimately chose to use a manually curated subset of MS~COCO, as it provides sufficient diversity and volume for our use case along with pixel-accurate segmentation masks. 
The manual annotation was conducted by the first author. To reduce classification noise, we employed the following exclusion criteria for the manually annotated subset of MS~COCO: (i) images depicting multiple people and (ii) images where gender representation was unclear to the annotators, due to occlusions, lighting conditions, or non-distinctive features. 
The resulting dataset contained images containing a single person in the foreground manually labeled as either male or female, which formed the basis for all subsequent experiments. The dataset comprises a total of 1,830 images, with 915 labeled as female and 915 as male. It was further divided into training (70\%), validation (15\%), and test sets (15\%). Specifically, the training set contains 1,198 images, while the validation and test sets each consist of 257 images. In our experiments, we used the MS~COCO ``person'' segmentation masks to define the spatially relevant region for a person depicted, with pixels inside the person mask treated as relevant ($0$ in ground truth mask \(A\)) and others as irrelevant ($1$ in \(A\)). It is important to note that these masks include the entire person including their clothing, which means that clothing-based bias is expected and represents part of the stereotypical associations we aim to address through XIL methods.

\subsection{Model Training \& Experimental Runs}

Each model was trained for 20 epochs with a learning rate of $1 \times 10^{-4}$ using the Adam optimizer. The performance was evaluated on the test set using metrics such as accuracy, loss (cross-entropy loss was used as the objective function), and explainability scores obtained using GradCAM and BLA explanations. For RRR, we utilized the person's mask from the MS~COCO dataset as the relevant area, setting these regions to 0 and the rest to 1. Ground truth segmentation masks are used to simulate user feedback, allowing for automated quantitative experiments. At each iteration of model steering with RRR, five samples are utilized using the chosen sampling strategy. 
For CAIPI we generated counterexamples by generating \(k\) copies of the selected image and applying image transformations on the irrelevant areas, as indicated by the ground truth mask from MS~COCO, while leaving the label unchanged.
At each iteration, five samples are selected using the chosen sampling strategy, and \(k\) counterexamples per sample are added to the initial training set, e.g., for \(k=3\), three counterexamples per sample are added ($15$ samples per iteration). For the hybrid approach, we provided \(5 \times k\) counterexamples per iteration to steer the model with RRR. For CAIPI and the hybrid approach we evaluated \(k=\{1,3,5\}\).

Each XIL approach (CAIPI, RRR, and hybrid) was evaluated across various settings. 
In addition to the six baseline models (no XIL steering), we conducted 22 further experiments with the best-performing baseline model, covering the three feedback integration approaches (CAIPI, RRR, and Hybrid), three  \(k\) values, two sampling strategies, and two explainability methods (GradCAM and BLA).


\subsection{Performance Metrics}
In addition to classification accuracy, the model's performance was evaluated using four metrics derived from the explanations of the applied explainability methods (for CAIPI we used GradCAM, and for RRR and the hybrid approach, the respective explainability method employed). To compute these metrics, GradCAM explanations were discretized by thresholding at the 25\% quantile, while BLA provided hard explanations directly.

\textbf{DICE score}  measures the alignment between the model's (binarized) attribution masks from the explainability method~(\(X\))  and the ground truth segmentation masks~(\(Y\)):
\begin{equation}
\text{DICE Score} = \frac{2 \cdot |X \cap Y|}{|X| + |Y|}
\label{eq:dice}
\end{equation}
The DICE score is a common segmentation metric that evaluates spatial overlap, making it suitable for our task. Higher scores indicate better alignment.

\textbf{Foreground Focus Proportion (FFP)} measures the proportion of foreground pixels with saliency values above a threshold $T$, indicating how much attention the model pays to the relevant foreground area in an image. Saliency values correspond to attributions produced by explainability methods.
\begin{equation}
\text{FFP} = \frac{\sum_{i \in F} u(S(i) > T)}{|F|}
\label{eq:ffp}
\end{equation}
where \(S(i)\) is the saliency at pixel \(i\), \(T\) is a threshold (25\% quantile of the saliency map), \(F\) is the foreground pixel set, and \(u(S(i) > T)\) is defined as $1$ if $S(i) > T$, and 0 otherwise.

\textbf{Background Focus Proportion (BFP)} is the proportion of background pixels with saliency values above a threshold:
\begin{equation}
\text{BFP} = \frac{\sum_{i \in B} u(S(i) > T)}{|B|}
\label{eq:bfp}
\end{equation}
where \(B\) is the background pixel set, and other terms follow the FFP definition above.

\textbf{Background Saliency Ratio (BSR)}  quantifies the proportion of total saliency attributed to the background:
\begin{equation}
\text{BSR} = \frac{\sum_{i \in B} S(i)}{\sum_{i \in I} S(i)}
\label{eq:bsr}
\end{equation}
where \(B\) is the background pixel set and \(I\) is the entire image.

\section{Results}

In this section, we present and discuss the results of the experiments, described in the experimental setup.

\begin{table*}[!t]
\renewcommand{\arraystretch}{1.3}
\caption{Comprehensive performance comparison of CAIPI, RRR, and the Hybrid approach: Bias metrics, Accuracy, and Misclassification Proportions. Bold values indicate the best performance for each metric.}
\label{table:combined_bias_metrics}
\centering
\begin{tabular}{c|c|c|c|c|c|c|c|c|c|c}
\hline
\textbf{Strategy} & \textbf{Sampling} & \textbf{XAI} & \textbf{\(k\)} & \textbf{FFP} $\uparrow$ & \textbf{BFP} $\downarrow$ & \textbf{BSR} $\downarrow$ & \textbf{DICE} $\uparrow$ & \textbf{Acc. (\%)} $\uparrow$ & \textbf{Miscl. Male (\%)} & \textbf{Miscl. Female (\%)} \\
\hline\hline
\begin{tabular}{c}Baseline \\ (no XIL)\end{tabular} & - & - & - & 0.35 & 0.31 & 0.74 & 0.316 & 74.58 & 46.2 & 53.8 \\
\hline
\multirow{6}{*}{CAIPI}
& Uncertainty & - & 1 & 0.34 & 0.22 & 0.70 & 0.343 & \textbf{75.42} & 47.2 & 52.8 \\
& Uncertainty & - & 3 & 0.38 & 0.26 & 0.73 & 0.386 & 72.03 & 46.7 & 53.3 \\
& Uncertainty & - & 5 & 0.22 & 0.31 & 0.82 & 0.229 & 71.19 & \textbf{51.3} & \textbf{48.7} \\
& High-conf. & - & 1 & 0.38 & 0.22 & 0.71 & 0.381 & 72.88 & 46.0 & 54.0 \\
& High-conf. & - & 3 & 0.42 & 0.23 & 0.65 & 0.422 & 64.41 & 47.1 & 52.9 \\
& High-conf. & - & 5 & \textbf{0.49} & 0.18 & 0.58 & \textbf{0.502} & 66.10 & 48.0 & 52.0 \\
\hline
\multirow{4}{*}{RRR}
& Uncertainty & GradCAM & - & 0.32 & 0.22 & 0.69 & 0.301 & 68.42 & 45.8 & 54.2 \\
& High-conf. & GradCAM & - & 0.36 & 0.20 & 0.67 & 0.328 & 71.12 & 47.4 & 52.6 \\
& Uncertainty & BLA & - & 0.35 & 0.23 & 0.66 & 0.338 & 69.85 & 46.1 & 53.9 \\
& High-conf. & BLA & - & 0.39 & 0.19 & 0.64 & 0.356 & 72.35 & 47.0 & 53.0 \\
\hline
\multirow{12}{*}{Hybrid}
& Uncertainty & GradCAM & 1 & 0.37 & 0.21 & 0.66 & 0.371 & 70.02 & 45.5 & 54.5 \\
& Uncertainty & GradCAM & 3 & 0.37 & 0.24 & 0.71 & 0.372 & 62.88 & 46.2 & 53.8 \\
& Uncertainty & GradCAM & 5 & 0.23 & 0.29 & 0.79 & 0.243 & 64.89 & 48.5 & 51.5 \\
& High-conf. & GradCAM & 1 & \textbf{0.49} & 0.19 & 0.51 & 0.487 & 73.32 & 47.1 & 52.9 \\
& High-conf. & GradCAM & 3 & 0.46 & 0.19 & 0.55 & 0.374 & 71.95 & 46.3 & 53.7 \\
& High-conf. & GradCAM & 5 & 0.43 & 0.21 & 0.52 & 0.429 & 70.85 & 47.0 & 53.0 \\
& Uncertainty & BLA & 1 & 0.40 & 0.20 & 0.63 & 0.395 & 71.58 & 45.2 & 54.8 \\
& Uncertainty & BLA & 3 & 0.39 & 0.22 & 0.68 & 0.398 & 64.15 & 46.5 & 53.5 \\
& Uncertainty & BLA & 5 & 0.26 & 0.27 & 0.76 & 0.278 & 66.42 & 48.0 & 52.0 \\
& High-conf. & BLA & 1 & 0.47 & 0.18 & 0.54 & 0.463 & 74.85 & 46.8 & 53.2 \\
& High-conf. & BLA & 3 & 0.48 & \textbf{0.17} & 0.52 & 0.412 & 73.22 & 46.5 & 53.5 \\
& High-conf. & BLA & 5 & 0.45 & 0.19 & \textbf{0.49} & 0.454 & 72.18 & 46.7 & 53.3 \\
\hline
\end{tabular}
\end{table*}

\subsection{Baseline Model Performance}

EfficientNet-B0 and VGG16 were the two best-performing models. We focus on reporting results for EfficientNet-B0 because it represents a more recent architecture.
In the baseline setting (without model steering), this model achieved an accuracy of 74,58\% and bias scores of 0.35 (FFP), 0.31 (BFP), 0.74 (BSR), and 0.316 (DICE). Additionally, GradCAM and BLA explanations were generated for the baseline model to assess the regions the model used for decision-making. 


\subsection{Comparison of Explanatory Interactive Learning Methods}

The evaluation results for all three investigated approaches are listed in Table~\ref{table:combined_bias_metrics}. 
The bias scores for the CAIPI-augmented models indicate improved focus on relevant foreground areas and reduced attention to irrelevant background regions compared to the baseline. These improvements became more pronounced as the value of 
\(k\) increased. CAIPI yields the best overall DICE score (using high confident samples). Regarding RRR, the bias scores show improved focus on relevant areas (lower BSR) compared to the baseline and the DICE scores reveal better performance for high confident samples, though the improvement for uncertain samples was less pronounced compared to CAIPI-augmented models. The hybrid approach resulted in the most effective focus on foreground regions, reducing reliance on irrelevant background areas as demonstrated by lower BFP and BSR values and a high FFP value.

Overall, the bias scores clearly show that the focus of the model and the data bias is reduced, especially by CAIPI and the hybrid approach compared to the baseline (e.g., increase of FFG from 0.35 to 0.49 for both CAIPI and the hybrid approach; drop of BFP from 0.31 to 0.18 for CAIPI; and drop of BSR from 0.74 to 0.51 for the hybrid approach). The effect of bias mitigation for RRR is less pronounced. At the same time, in most cases classification accuracy drops by a few percentage points compared to the baseline, which demonstrates the costs of improved explainability and  bias mitigation. An exception, however, is CAIPI (with uncertain samples and k=1) where accuracy slightly improves (from 74.58\%  to 75.42\%). This demonstrates that improved model quality not necessarily has to be at the cost of accuracy. 

\begin{figure*}[tb]
    \centering
    \includegraphics[width=1\linewidth]{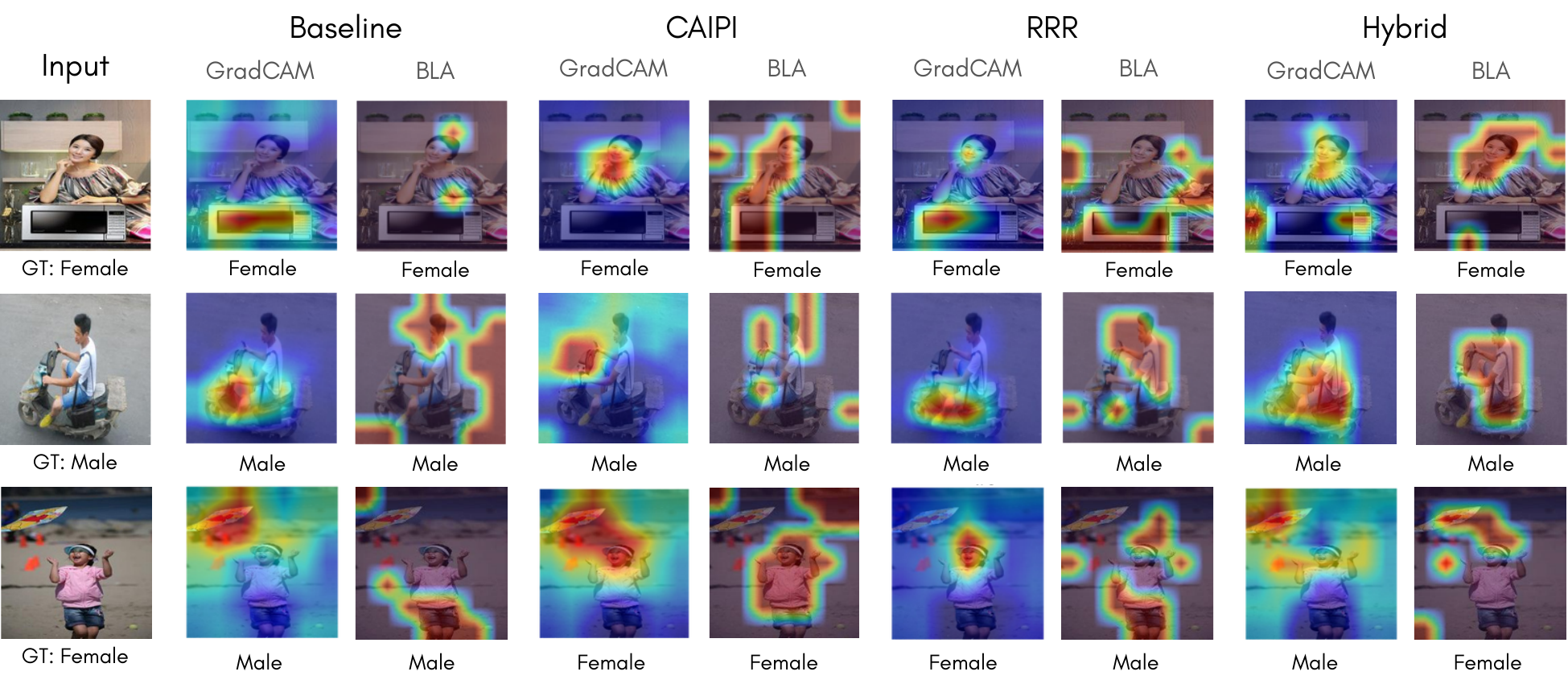}
    \caption{GradCAM and BLA explanations for EfficientNet-B0 predictions (predicted class below subfigure) before (Baseline) and after XIL steering.}
    \label{fig:beforeafter}
\end{figure*}

Sampling strategy and choice of explanation method had a substantial impact on the outcomes. In most cases, high-confidence sampling outperformed uncertainty sampling, suggesting that more reliable model predictions provide a stronger basis for effective feedback integration. Similarly, the comparison between GradCAM and BLA revealed slight differences in classification accuracy, with BLA yielding better results in most cases. Visually, we can also observe clear differences between the two methods (Fig.~\ref{fig:beforeafter}), which can be attributed to their distinct methodologies. GradCAM is a post-hoc explainability approach applied after model training, whereas BLA is a self-explaining method that is jointly learned with the model. These findings highlight the sensitivity of XIL approaches to both the sampling strategy and the underlying XAI method.

\subsection{Analysis of Gender Bias and Mitigation Effects}

To assess fairness~\cite{mehrabi2021survey}, we analyzed the proportion of male and female misclassifications across various experimental setups. Table~\ref{table:combined_bias_metrics} reveals a general trend of females being misclassified more often than males (baseline: 46.2\% vs. 53.8\%). Overall, most  approaches achieved a more balanced distribution of misclassifications (e.g., CAIPI with high-confidence sampling and $k=5$: 48.0\% vs. 52.0\%), demonstrating their potential to improve the fairness of the models. 

\subsection{Limitations} In our study, user interactions were simulated, so evaluating the methods with human users remains an important next step. Our XIL evaluation relies on precise segmentations of the input images, thus in practical settings, this requirement would need to be relaxed, e.g., by using bounding boxes or coarse annotations. Furthermore, all experiments were performed on a single, manually curated dataset, which offers controlled conditions but limits the generalizability of the findings.

\section{Conclusion}


The key findings from our experiments are: (i) The investigated XIL methods improve the model's focus on relevant regions, but this effect is much more pronounced in CAIPI and our hybrid strategy than in RRR (Fig.~\ref{fig:beforeafter}); (ii) The proposed hybrid strategy shows strong bias mitigation potential and combines mutual strengths of CAIPI and RRR; (iii) High-confidence sampling is more useful than uncertainty sampling for XIL (opposite to what is observed in Active Learning).

Our experiments demonstrate the effectiveness of XIL for bias mitigation. Furthermore, integrating CAIPI or our hybrid approach guides the models to focus more on relevant regions and less on irrelevant regions in the images. The analysis of gender-specific misclassifications further highlights the potential of feedback integration strategies to improve the fairness of classifiers.
%
Future research should assess the performance of XIL with human users in a dedicated user study as well as explore XIL for multi-label and multi-class problems.


\section*{Acknowledgements}
This research was funded by the Austrian Science Fund (FWF) \href{https://doi.org/10.55776/P36453}{10.55776/P36453} and \href{https://doi.org/10.55776/COE12}{10.55776/COE12}, as well as  the Austrian Promotion Agency (FFG) under project grants 898085 (TrustAI) and FO999904624 (FairAI).

\clearpage
\bibliographystyle{IEEEtran}
\bibliography{bibliography2}

\end{document}